# Brain-inspired spike-timing plasticity for reliable label-efficient event-camera vision

Mohamad Yazan Sadoun, Sarah Sharif, Yaser Mike Banad*

*School of Electrical and Computer Engineering, University of Oklahoma, Norman, OK, USA*

*Corresponding author: bana@ou.edu

## Abstract

Deploying event-camera object detectors is gated by per-frame label budgets and GPU compute. Three local spike-timing-dependent plasticity (STDP) modules (sequence, candidate, and tube reliability) close both on a single CPU thread, no GPU. On the FRED drone benchmark, three supervision tiers span the label-efficient regime: a strict zero-label detector reaches 53.8%; ≈26 train-derived bits give 76.9%; an STDP candidate-reliability gate reaches 78.60±0.42% mAP@30. Under acquisition-order drift the cohort gate beats streaming k-means by +2.03 ± 0.58 pp (20/20 positive); a no-drift control falsifies the effect. STDP tightens single-model variance 6.6×; one trained gate matches a 44-seed ensemble bound. The gate ports to Intel Lava at 89% top-2 agreement; on EV-UAV, a tube-level STDP layer cuts false alarms from 454 to 331 × $10^{-4}$ at $P_d$ ≥ 88%. Dense gradient-trained detectors cannot deliver this combination by construction: gradient training, dense matmul, and no local plasticity rule.

Biological retinas spike when the world changes; silicon event cameras [1, 2] share that operating principle, emitting asynchronous brightness-change events at microsecond resolution instead of frames at fixed cadence. Spiking neural networks (SNNs) process those binary streams through local, gradient-free Hebbian plasticity rules [3, 4], including recent online SNN estimation and control demonstrations [5, 6] and bio-inspired multi-UAV perception [7]. The pairing of an event-driven sensor with event-driven compute defines neuromorphic vision: a perception stack that targets microsecond latency and watt-scale power on commodity neuromorphic substrates and the spiking circuit primitives underneath them [8–15].

In practice, the most accurate event-camera object detectors today are dense, gradient-trained convolutional, recurrent, transformer, or multimodal fusion networks [16–21], each consuming large labelled datasets during training and a GPU-class inference path. Parallel event-only work shows that drone-specific frequency structure can outperform frame-style detectors when exploited directly [22]. Deploying these detectors is gated on per-target annotation budgets and on GPU availability at inference.

A growing line of work [23–26] closes this gap by replacing the dense detector with a spiking architecture trained end-to-end. End-to-end spike-driven detectors at competitive accuracy use millions to tens of millions of synapses, span many cores and chips on Loihi-2 and SpiNNaker [8, 12], and require surrogate-gradient training on GPU with full label budgets.

Here we pair six classical event-only detection channels with three local STDP modules acting at the sequence, candidate, and tube levels. At the sequence level, a 384-plastic-synapse spiking gate selects one of six processing recipes per sequence from an event-statistics fingerprint, analogous to mixture-of-experts routing [27, 28]. At the candidate level, a reward-modulated STDP candidate-reliability gate neuromodulates anchor confidences and admits temporally coherent rescues. At the tube level, STDP-Tube links detections into short event tubes and learns tube reliability from temporal continuity, event support, and cross-channel evidence. The sequence gate is trained gradient-free by spike-timing-dependent plasticity (STDP) [3, 4, 29–31], converting broad random-projection weights into sparse attractor weights; with those weights frozen at deployment the gate fits within a single neuromorphic core, runs on one CPU thread, and is stable under simulated 8-bit Loihi-2 quantisation. The term label-efficient, used throughout, has the operational meaning given in Methods §Supervision footprint.

We evaluate on FRED [16] and EV-UAV [32] event-camera drone benchmarks. Three results anchor the paper. First, a Phase-1 STDP-trained sequence gate's single-model variance meets the analytic $N^*=44$ random-seed sample-mean bound under the streaming deployment protocol, and a matched no-drift control isolates drift-tracking as the mechanism behind the gate's +2.03 ± 0.58 pp gain under acquisition-order shift. Second, FRED reports three explicit supervision tiers (53.8%/76.9%/78.60% at 0/≈26/≥3,072 stored deployment bits). Third, STDP-Tube improves the EV-UAV operating point at the candidate level, and the sequence cohort gate ports to Intel Lava at 89.4% (42 of 47) top-2 cohort preservation under simulated 8-bit Loihi-2 precision on a single CPU thread.

# Related work

***Event-camera object detectors.***

The strongest event-camera detectors are gradient-trained dense networks: RVT [17] couples a recurrent backbone to a transformer-style head and reports 47.2 mAP on Gen1; HMNet [18] uses hierarchical memory; DAGr [19] fuses asynchronous graph and image streams; AhMNet [21] aggregates heterogeneous attention across modalities; ReYOLOv8 [20] adapts YOLOv8 to event tensors with temporal recurrence. FRED [16] releases the benchmark and reports dense-detector baselines on the same canonical split we use. These detectors absorb large per-frame label budgets during training and run on GPU at inference. DDHF [22] shows event-only frequency signatures outperform frame-style detectors on drone targets, motivating our rotor-frequency channel.

***End-to-end spiking detectors and neuromorphic substrates.***

A parallel line replaces the dense backbone with a fully spiking network: EMS-YOLO [24] (ResNet-18 / ResNet-34 spiking backbones at 9.3/14.4 M parameters), Spike-YOLO [25], and EAS-SNN [26] (YOLOX-S / YOLOX-L spiking variants at 8.9/25.3 M parameters) all train end-to-end with surrogate gradients on GPU, inheriting the label and compute footprint they were intended to escape. Loihi-2 [8, 10, 11] and SpiNNaker [12] are the dominant inference substrates these systems target; smaller-scale demonstrations report watt-scale inference power for keyword spotting [33]. Recent work in our group [5–7, 13–15] addresses adjacent neuromorphic deployment, device-physics, and bio-inspired UAV-perception questions.

***STDP classifiers, routing, and test-time adaptation.***

The cohort gate's training rule is in the lineage of unsupervised STDP-with-WTA classification [29–31], with biological grounding in Bi-Poo [3], Turrigiano synaptic scaling [34], Triesch intrinsic plasticity [35], and Maass WTA cortical competition [36]. The cohort-to-recipe lookup is functionally analogous to gradient-free test-time adaptation methods [37–39] but emits a discrete recipe choice per sequence with $\log_2|L|$ bits per non-empty cohort, and to sparse mixture-of-experts routing [27, 28] but at a single-core synapse count.

# Results

The pipeline carries three local STDP modules at different levels (Tab. 1); subsequent subsections attribute each empirical claim to the relevant module.

**Table 1:** The three local-plasticity modules. Two carry 384 plastic synapses by design parity. Subsequent tables and figures use these module names.

| Module | Dataset / level | Architecture | Plastic synapses | Owns which claim |
|---|---|---|---|---|
| Sequence cohort gate | FRED, sequence routing | 17→32→12 LIF, hard WTA, plastic Who | 32×12 = 384 | Drift +2.03±0.58 pp; 6.6× variance tightening. |
| R-STDP candidate-reliability gate | FRED, candidate level | 64 hidden WTA → 6 reliability states | 64×6 = 384 | 78.60±0.42% mAP@30 (N=10 seeds). |
| STDP-Tube | EV-UAV, tube level | bounded WTA/STDP → 6 states | locked K=3, 5 | Fa=$331\times10^{-4}$ at Pd≥88% (vs 454 baseline). |

## Label-free and low-label detection tracks on FRED

We compose six classical event-based detection channels under an IoU-distinctness union rule (Fig. 1): a density-watershed clusterer, a multi-hypothesis K-means tracker, a temporal-pooling channel, a rotor-frequency analyser, a polarity-asymmetry filter, and a shadow-aware rescoring stage. The union accepts every channel's top-1 candidate per frame and appends auxiliary candidates whose intersection-over-union with the running set falls below a fixed train-derived threshold. The first five channels form the strict label-free detector: their parameters are fixed from event statistics or sensor geometry and use no ground-truth boxes. The sixth shadow-aware stage and the union threshold define the train-calibrated track: they use training labels once to set a compact calibration, then consume no labels at inference.

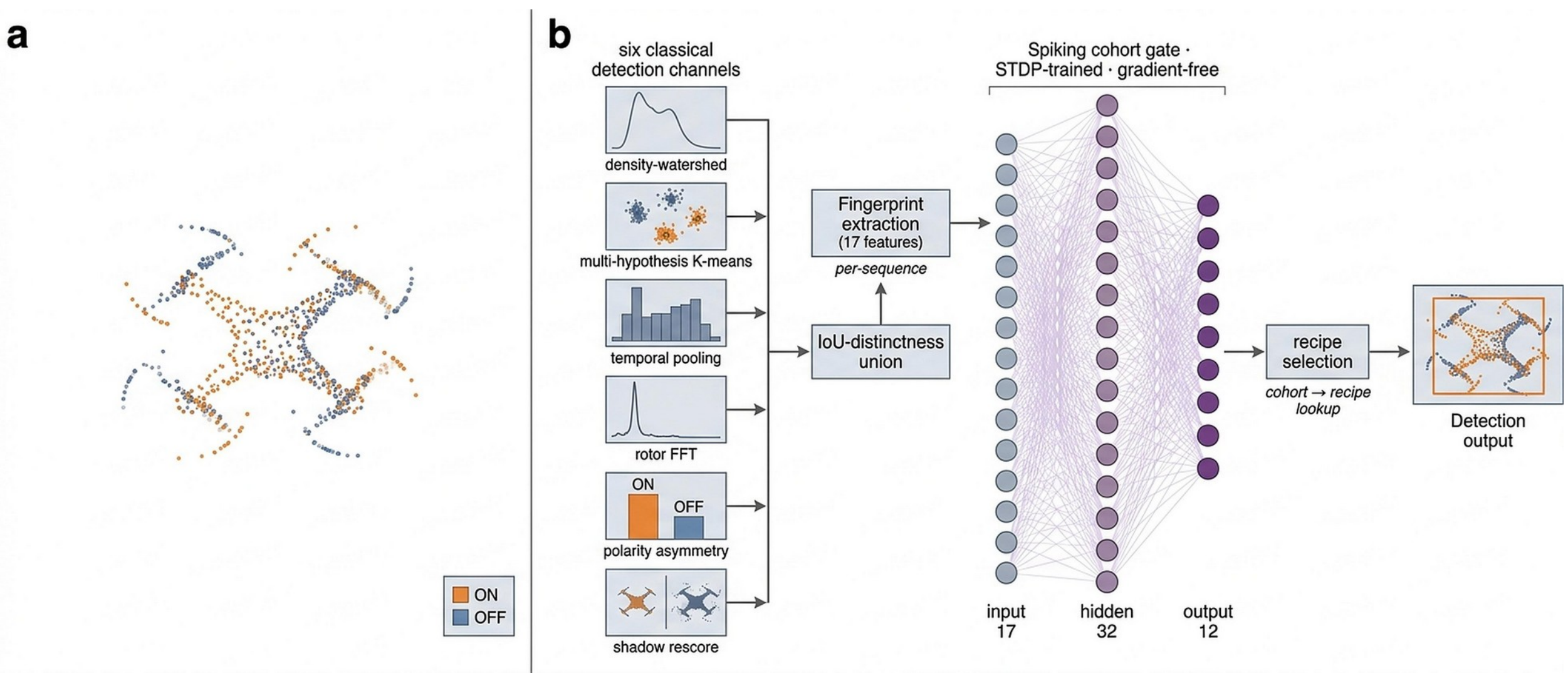

**Fig. 1:** Pipeline overview. **a,** ON/OFF event scatter for a drone in flight. **b,** Six classical event-detection channels feed an IoU-distinctness union; a 17-dimensional fingerprint drives a 384-plastic-synapse STDP-trained spiking cohort gate (17 → 32 → 12 LIF) that selects one of six processing recipes per cohort.

On the canonical FRED split (184 train, 47 test, multi-drone ground truth rebuilt from raw acquisition logs), we report three supervision tiers (Tab. 2; Supplementary Tab. S2). The strict label-free five-channel detector uses no train labels and reaches 53.8% test mAP@30, 31.4% mAP@50. The support-aware six-channel detector adds an IoU-distinctness threshold $\tau \in \{0.2, 0.3, 0.5\}$ and four shadow-rescore linear weights (≈26 train-derived bits) and reaches 76.9% mAP@30 (95% sequence-level bootstrap CI [40] [71.1, 82.2]), 51.7% mAP@50; the support-aware confidence term is label-free at inference. A competition-aware reward-modulated STDP candidate-reliability gate trained on the same split (locked policy: preserve six-channel anchors, modulate confidence, add only temporally coherent non-anchor candidates) reaches 78.60 ± 0.42% mAP@30 and 52.57 ± 0.43% mAP@50 (mean ± s.d. over N=10 seeds; range [78.02, 79.34]; all 10 above the support-aware reference; locked seed-23 checkpoint at 78.14%/52.67%, footnote d). A six-channel candidate oracle covers 86.6% of test targets at IoU 0.30 and 72.2% at IoU 0.50. For deployment-context comparison, the dataset paper [16] reports its strongest dense event-only baseline (YOLOv11, full per-frame box labels, GPU inference) at 87.68% mAP@50 and 49.25% mAP@50:95 on the same canonical test split; this paper's three supervision tiers occupy a different operating point in the supervision-cost / compute / accuracy trade-off space (Supplementary Tab. S2).

## Online spike-timing plasticity adapts to distribution shift

A 384-plastic-synapse leaky integrate-and-fire (LIF) cohort gate above the six channels selects which processing recipe to apply per sequence (Fig. 1b), trained gradient-free by STDP on a 17-dimensional event-statistic fingerprint. Splitting training fingerprints by acquisition order into initial/shifted streams and presenting shifted-stream sequences one at a time (Fig. 2), the deployed frozen-Who configuration delivers +2.03 ± 0.58 pp mean-stream-mAP@30 over streaming k-means at N=20 seeds, with 20/20 seeds positive (range +1.06 to +2.95; Tab. 3 row 3); an online-update variant gives +1.90 ± 0.62 pp (matched N=19: +1.92 ± 0.63 pp). At the same train-derived supervision footprint as the hand-tuned cascade (Tab. 2), the spiking cohort gate reduces the across-seed standard deviation of mean-stream-mAP@30 by 6.6× ($\sigma$ = 0.607 → 0.092 pp; Tab. 3 rows 1 vs 3): a single trained gate matches the analytic sample-mean variance bound of a 44-seed random-init ensemble (Methods). On the predeclared four-recipe pool $L_4$, the gate's per-sequence test mAP@30 is TOST-equivalent at ±0.5 pp to five established cohort baselines (k-means, GMM, MLP, LAME, and SHOT-IM; $p_{TOST} \leq 2.2\times10^{-3}$; Methods §Statistical analysis), framing the gate's contribution as drift adaptation and variance tightening at parity accuracy. A random-50/50 no-drift control returns the advantage to −0.44 ± 1.97 pp (6/20 positive; Fig. 2d), with frozen k-means rising to 74.21%: the advantage is positive when, and only when, drift is present. Per-sequence retention is 28/30 within ±0.5 pp of initial-gate performance, with one sequence improving by +0.94 pp and one regressing by −2.76 pp (test ID 111).

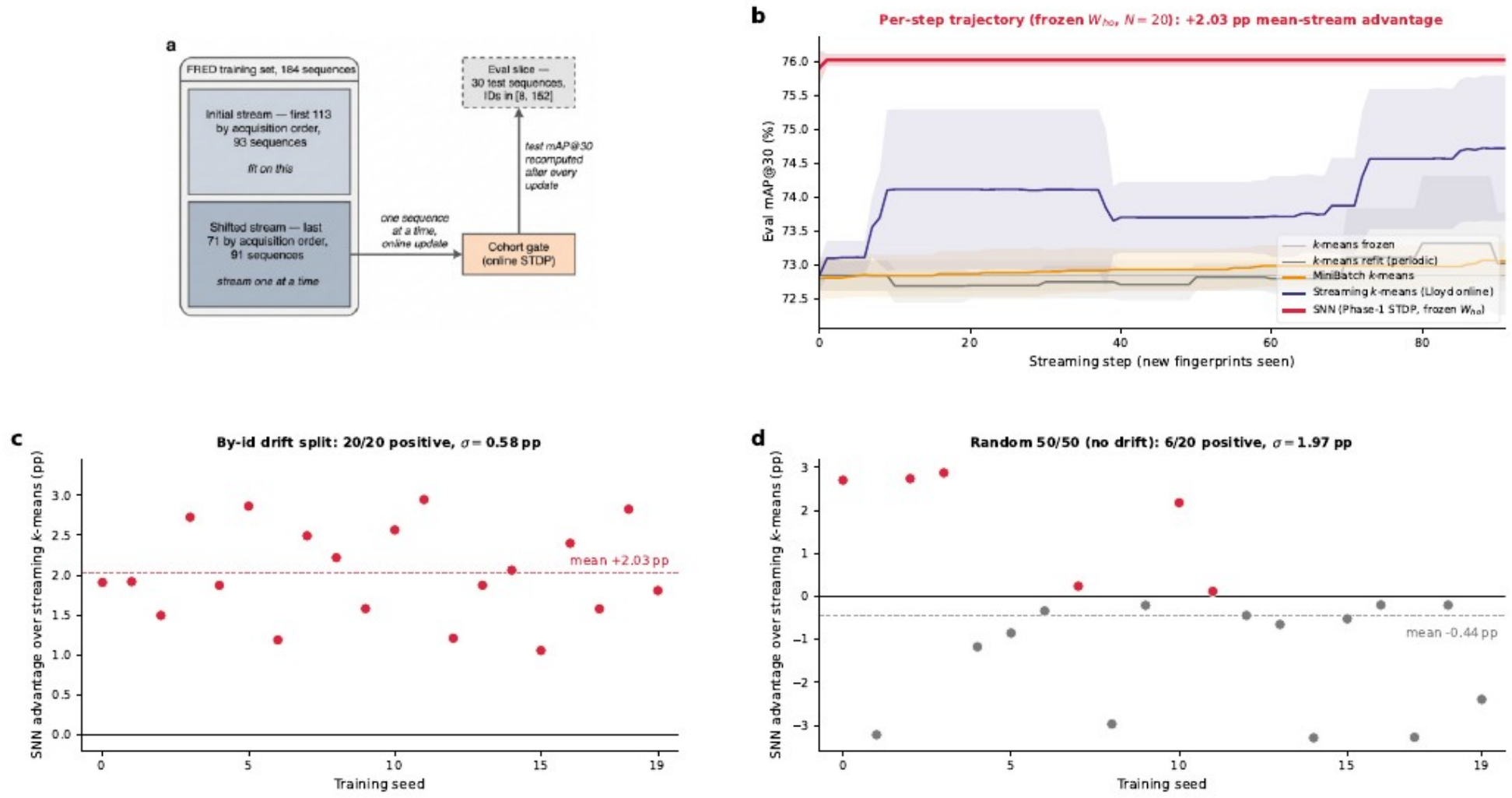

**Fig. 2:** Streaming distribution-shift adaptation with a no-drift control. **a,** Streaming protocol: 184 FRED training sequences split by acquisition order into initial (n=113) and shifted (n=71) streams; the cohort gate is reassigned for a held-out 30-sequence evaluation slice after every update. **b,** Per-step Eval mAP@30, mean ± std across N=20 frozen-Who seeds, five methods. **c,** Per-seed SNN advantage over streaming k-means under the by-id drift split: +2.03 ± 0.58 pp, 20/20 positive. **d,** No-drift control on a random 50/50 split: −0.44 ± 1.97 pp, 6/20 positive; the advantage is present if and only if drift is.

## Mechanism decomposition: architecture and plasticity together deliver drift adaptation with deployment-grade reproducibility

A four-condition ablation (random init vs Phase-1 STDP, × frozen vs online $W_{ho}$ during streaming, N=19 matched seeds; Tab. 3) isolates two complementary deployment-relevant contributions inside the spiking gate. The spike-coded random-projection architecture contributes a mean drift shift of +1.86 ± 0.96 pp (18/19 positive). Phase-1 STDP plasticity contributes a 6.6× tightening of seed-to-seed standard deviation (σ = 0.607 → 0.092 pp; these are the streaming-protocol σ values used for N*=44) plus a small mean refinement (+0.18 ± 0.62 pp paired; 8/19 positive). Online STDP during streaming is statistically equivalent to Phase-1-frozen (Δ = −0.13 ± 0.22 pp paired). Architecture and plasticity contribute complementary deployment-relevant properties (a mean shift and a variance contraction respectively), both inside the spiking compute fabric; together they give the deployed gate the +2.03 ± 0.58 pp drift advantage at single-model variance σ = 0.092 pp reported in row 3 of Tab. 3.

### *Mechanism of the 6.6× tightening.*

Phase-1 STDP drops the per-cohort $W_{ho}$-column entropy from 3.24 to 2.30 nats (−29%) and lifts the hidden-unit row-norm coefficient of variation (CV) from 0.133 to 1.43 (10.8×, unrounded 0.1328 → 1.4281), concentrating each cohort's drive on a small subset of hidden units (Fig. 3). Per-seed matrices remain seed-specific (column-entropy seed-spread *rises*, 0.013 → 0.289 nats) yet operational cohort routing converges to seed-invariant assignments — the signature of STDP-with-winner-take-all (WTA) attractor classification [30, 31].

## Single-model variance reaches the analytic 44-seed sample-mean bound

From a random-init pool the sample-mean variability is bounded by $\sigma_{rand}/\sqrt{N}$; the smallest N at which this meets the Phase-1 STDP single-model standard deviation $\sigma_{STDP}$ = 0.092 pp is $N^* = \lceil(\sigma_{rand}/\sigma_{STDP})^2\rceil$ = 44 seeds on the streaming mean-stream-mAP@30 metric (Fig. 3; $\sigma_{rand}$ = 0.607 pp from the architecture-only ablation, Tab. 3 at N=19 matched seeds). An empirical plurality-vote ensemble of independently random-initialised gates approaches this analytic bound within 1.45× at N=44 (Methods + §S7); the analytic bound is the operationally relevant comparator because it characterises the trained single-model's deployment variance, not a deployment-time ensemble.

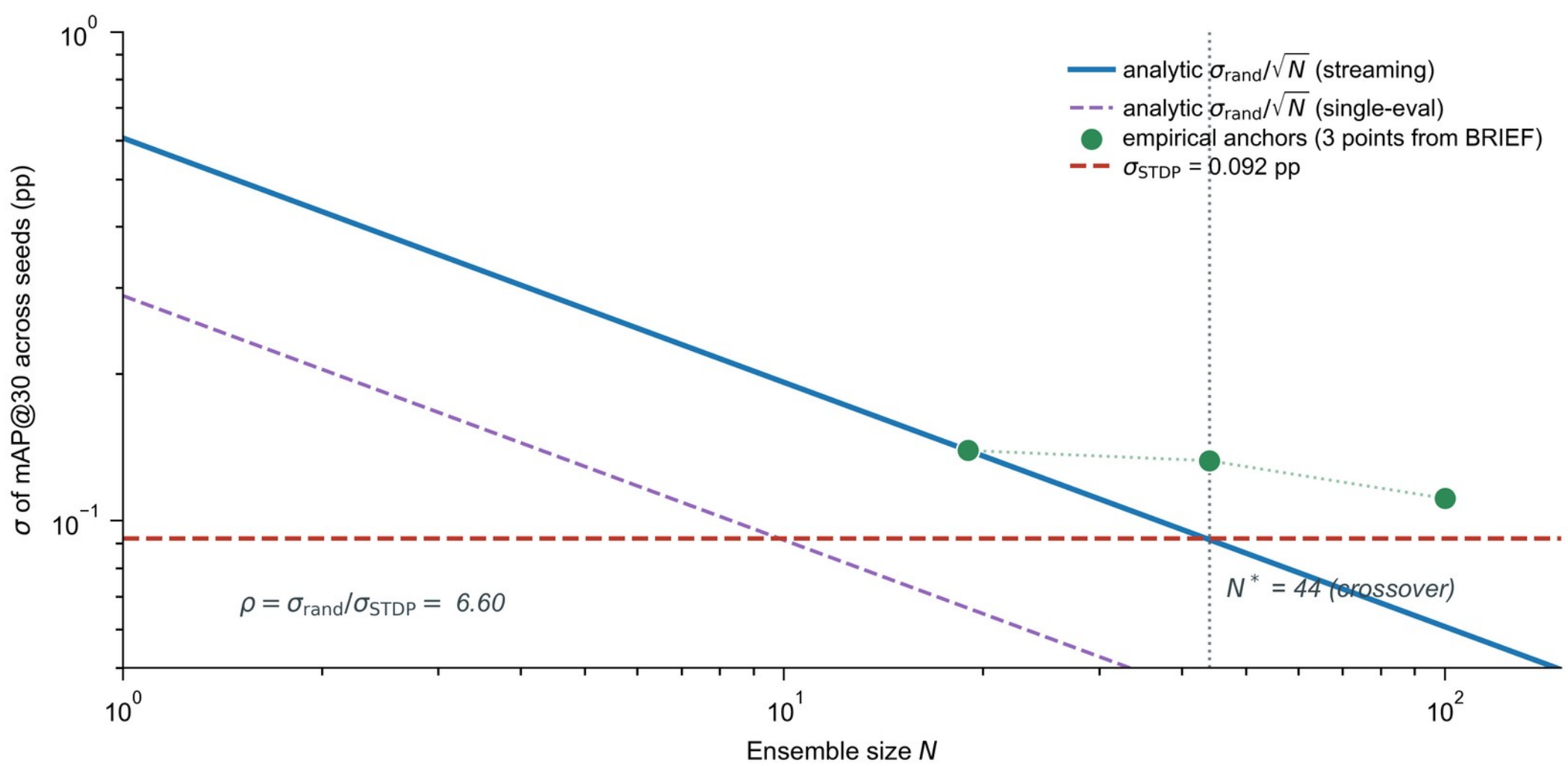

**Fig. 3:** Log-log plot of mAP@30 standard deviation versus ensemble size N. The streaming-protocol analytic bound $\sigma_{rand}/\sqrt{N}$ (with $\sigma_{rand}$=0.607 pp from N=19 architecture-only seeds; Tab. 3) intersects $\sigma_{STDP}$=0.092 pp at N*=44. Empirical plurality-vote anchors approach this bound within 1.45× at N=44.

## STDP-Tube improves EV-UAV operating points

EV-UAV [32] uses DAVIS346 sensors with few-pixel targets and reports event-level IoU, $P_d$, and $F_a$. The six-channel baseline with sensor-geometric rescaling and a label-free hot-pixel pre-filter over-emits on test: $P_d$ = 91.17%, $F_a$ = 662.34 × $10^{-4}$, event IoU 31.18% (Tab. 4). STDP-Tube links detections into short event tubes and applies bounded local STDP to tube features (temporal continuity, support density, cross-channel agreement, isolation; Fig. 4a). K=3 (balanced) and K=5 (low-$F_a$) are locked on full validation before testing. On test, K=5 improves mAP@30 from 50.97% to 57.03%, event IoU 31.18% to 44.48%, and $F_a$ from 662.34 to 340.21 × $10^{-4}$ while retaining $P_d$ = 88.44%; K=3 gives the strongest mAP@50 (17.92% vs 16.44% baseline). Score-threshold curves (Fig. 4b) at $P_d$ ≥ 88% give $F_a$ = 331.16 for STDP-Tube K=5 vs 454.16 baseline and 364.37 fixed-length-5 tube filter.

## Substrate, quantisation, and CPU-thread deployment

The trained gate ports to the Intel Lava neuromorphic simulator with matched LIF dynamics (Supplementary Tab. S3): 89.4% top-2 cohort preservation on the full 47-sequence canonical test split. Full 8-bit fixed-point quantisation at Loihi-2 precision (every weight, membrane register, STDP trace, homeostatic threshold; PyTorch simulation) keeps test mAP@30 within the seed-noise floor in the quantisation probe (+0.39 pp relative to fp32), while the 6-bit sanity check degrades by −1.92 pp. End-to-end inference runs on a single CPU thread at ~20 ms per frame compute-side (gate adds ~0.04 ms amortised), matching the pipeline's 30 Hz cadence; the prototype's ~47 ms-per-frame end-to-end runtime is preprocessing-I/O bound. The gate carries 384 plastic + ≈272 fixed-random projection synapses (≈656 total), within current Loihi-2 and SpiNNaker per-core capacity, with an itemised Loihi-1 energy proxy of ~83 µJ per cohort classification. No GPU, no accelerator. Predictions are tight across drone types on six held-out sequences (Fig. 5).

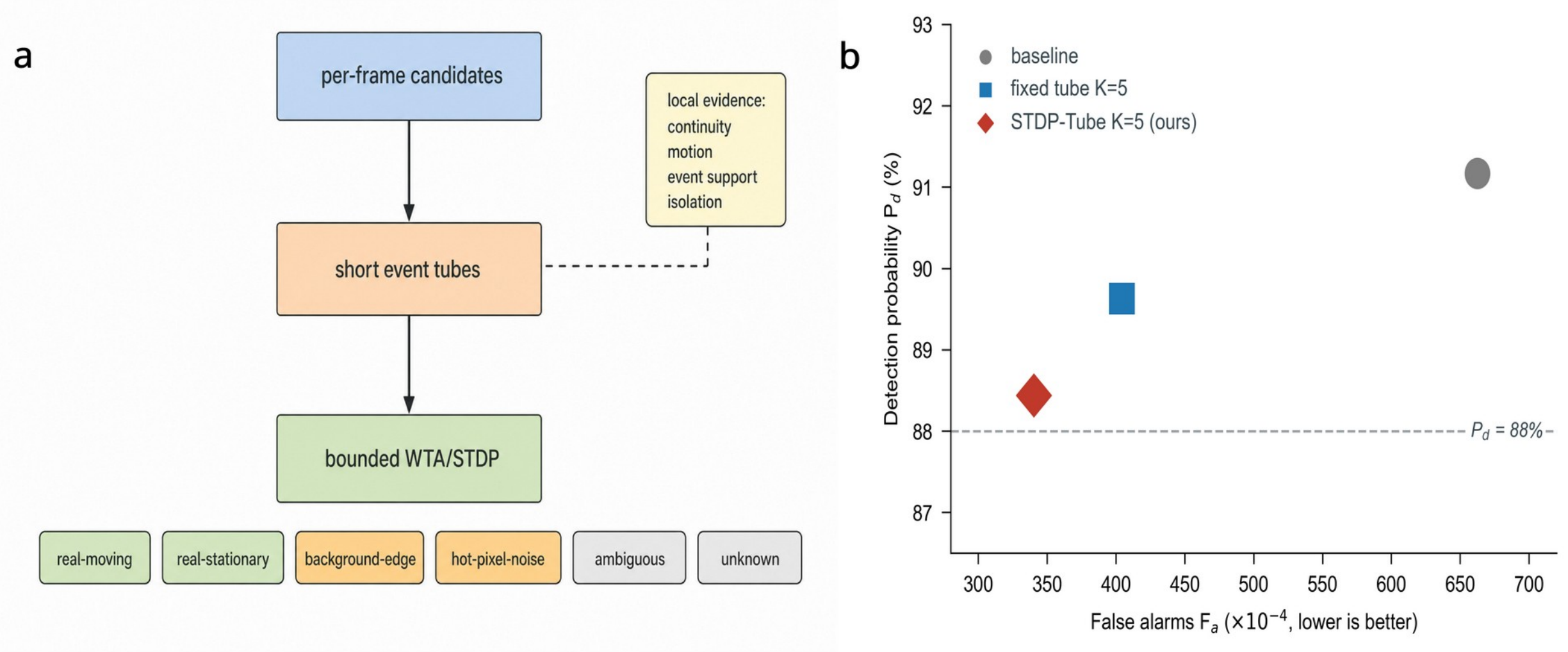

**Fig. 4:** Candidate-level reliability with STDP-Tube on EV-UAV. **a,** STDP-Tube links per-frame candidates into short event tubes and maps tube features to six output states (including an explicit *unknown*) via a bounded WTA/STDP reliability layer. **b,** EV-UAV operating points at locked defaults (Tab. 4): STDP-Tube K=5 (diamond) reaches $F_a$ = 340.21×10$^{-4}$ at $P_d$ = 88.44%, below the six-channel baseline ($F_a$ = 662.34) and the fixed-length-5 tube filter; dashed line marks $P_d$ = 88%.

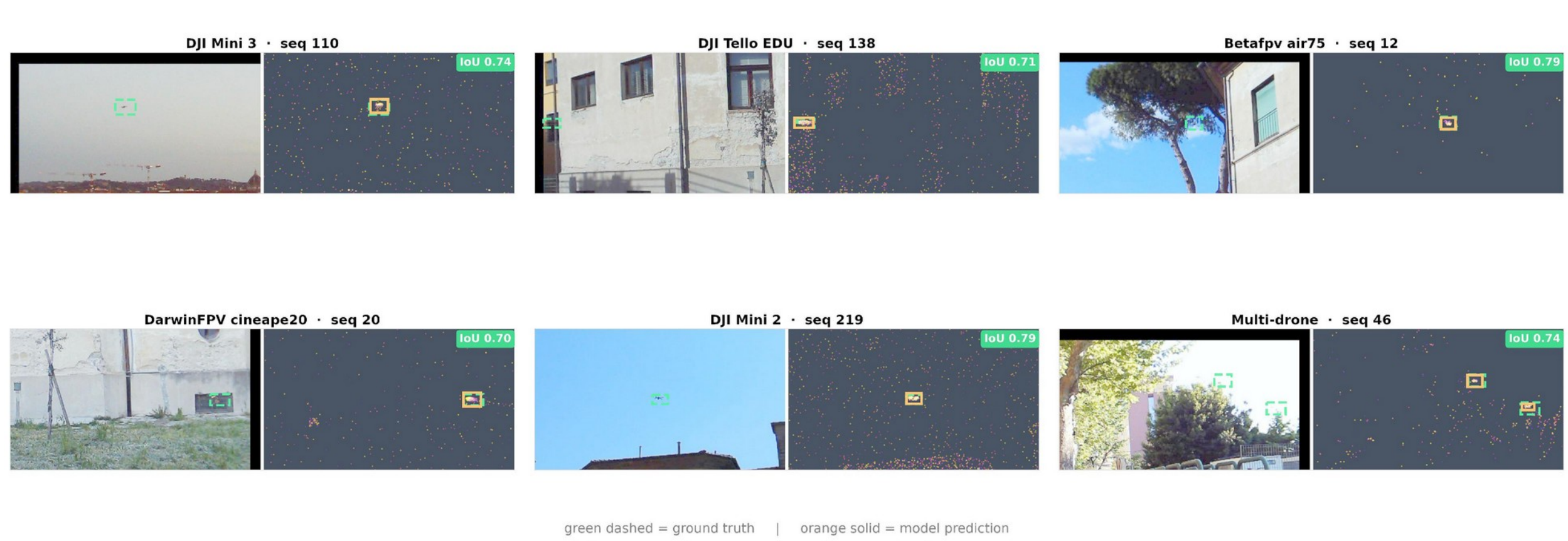

**Fig. 5:** Six representative sequences across five drone types (DJI Mini 3, DJI Tello EDU, DJI Mini 2, BetaFPV Air75, DarwinFPV CineApe 20) plus a multi-drone scene. RGB ground-truth view and event-stream view shown side by side at the same instant; green dashed box = ground truth, orange solid = model prediction.

**Table 2:** Detection on the FRED canonical split, by supervision tier. SNN rows: mean ± s.d. over N=20 seeds, except R-STDP test row at N=10 (footnote d). Brackets give 95% sequence-level bootstrap CI. Dense reference (YOLOv11 event) is from [16] Table 2 for context; their reported metric is mAP@50:95.

| Method | Label use | Train mAP@30 | Test mAP@30 | Test mAP@50 |
|---|---|---|---|---|

| **Ours (this paper)** | | | | |
|---|---|---|---|---|
| Five-channel event union | none | 54.49 | 53.81 | 31.43 |
| STDP cohort gate, label-free | none | 79.78±0.13 | 75.09±0.58 | 49.74±0.85 |
| Six-channel support confidence | compact train calibration [a] | 81.63 | 76.87 [71.1, 82.2] | 51.67 |
| Six-channel + R-STDP candidate reliability | train-calibrated STDP [d] | 82.18 | 78.60±0.42 | 52.57±0.43 |
| **Classical baselines (no neural training, no per-frame labels)** | | | | |
| DBSCAN + centroid box | none | — | 21.02 | 6.85 |
| ATCC + temporal track | none | — | 34.29 | 12.91 |
| Density watershed, one channel | none | 68.54 | 65.94 | 40.61 |
| Multi-hypothesis k-means, one channel | none | — | 46.27 | 30.08 |
| Six-channel deterministic fusion | none | 79.22 | 72.47 | 45.21 |
| Hand-tuned cascade | train tuned | 81.22 | 75.37 | 50.21 |
| **Dense supervised reference / Oracle** | | | | |
| YOLOv11 (event) [16] | per-frame boxes | — | — (87.68 at mAP@50) | 49.25 mAP@50:95 |
| Six-channel candidate oracle [c] | post-hoc GT diagnostic | 89.69 | 86.60 | 72.17 |

[a] Compact train calibration: three-value train-mAP sweep for τ plus four shadow-rescore weights (~26 stored bits). [c] Oracle uses ground truth only after candidate generation to ask whether any six-channel candidate overlaps each target at the requested IoU threshold; coverage, not AP. [d] Competition-aware reward-modulated STDP gate with frame-local WTA confidence modulation and temporally coherent non-anchor candidate admission; train labels modulate plastic reliability weights; canonical test emitted once. Test row mean ± s.d. over N=10 seeds. The locked seed-23 checkpoint reports 78.14 / 52.67 on canonical test and reproduces to within 0.01 pp.

**Table 3:** Mechanism decomposition at N=19 matched seeds. Top block: random init vs. Phase-1 STDP, frozen vs. online deployment; the frozen-Who row anchors N*=44. Middle/bottom: spike encoding vs. plasticity rule vs. rate-coded controls.

| Tested mechanism | Stream mAP@30 (%) | Seed s.d. (pp) | Gain over stream k-means (pp) | p Holm [b] |
|---|---|---|---|---|
| **Architecture × plasticity ablation (spike-coded)** | | | | |
| Random init, frozen Who | 75.83 ± 0.61 | 0.607 | +1.86 ± 0.96 (18/19) | — |
| Random init, online STDP | 75.90 ± 0.25 | 0.253 | +1.93 ± 0.64 (19/19) | — |
| Phase-1 STDP, frozen Who deployment | 76.01 ± 0.09 | 0.092 | +2.04 ± 0.60 (19/19) | — |

| Phase-1 STDP, online STDP [d] | 75.88 ± 0.22 | 0.220 | +1.92 ± 0.63 (19/19) | $5\times10^{-5}$ |
|---|---|---|---|---|
| **Alternative plasticity rules (spike-coded)** | | | | |
| Spike + η-matched STDP [a] | — | 0.24 | +1.92 ± 0.66 (19/19) | $5\times10^{-5}$ |
| Spike + R-STDP [a] | — | 0.52 | +1.80 ± 0.76 (18/19) | $8\times10^{-5}$ |
| Spike + online k-means | — | — [c] | +2.09 ± 0.63 (19/19) | — |
| **Rate-coded controls (no spikes)** | | | | |
| Rate + online k-means | — | 1.06 | +0.30 ± 1.31 (10/19) | — |
| Rate + frozen centroids | — | 1.39 | −0.14 ± 1.51 (8/19) | — |
| **Reference baselines** | | | | |
| Streaming k-means (α=0.10) | 73.98 ± 0.62 | — | 0 | — |
| Frozen k-means | 72.84 ± 0.22 | — | −1.14 | — |

[a] Pairwise-equivalent to headline at p_Holm = 1.00. [b] vs. rate+k-means. [c] k-means label-free routing collapses to a single recipe; per-seed mAP invariant by construction. [d] Online-streaming row; deployment configuration (Who frozen, row 3) anchors N*=44. Extension to N=20 on row 3: 76.01±0.09%, σ=0.090 pp, +2.03±0.58 pp (20/20) — the abstract's headline drift number.

**Table 4:** EV-UAV full-test detection metrics. Our rows are event-only label-free variants run on the same 24 test sequences; supervised references are copied from Table 2 of ref. [32] and use full event-segmentation labels. Our mAP columns use boxes derived from event masks at 20 ms accumulation; the copied supervised rows do not report box mAP. $P_d$ and event IoU are higher-better; false alarms are lower-better and reported at $10^{-4}$ scale.

| Method | Label use | mAP@30 | mAP@50 | Pd | False alarms / event IoU |
|---|---|---|---|---|---|
| **Ours, label-free event detection** | | | | | |
| Six-channel baseline | none | 50.97 | 16.44 | 91.17 | 662.34 / 31.18 |
| Fixed event-tube filter, K=3 | none | 55.95 | 17.46 | 90.38 | 487.94 / 37.39 |
| Fixed event-tube filter, K=5 | none | 56.26 | 17.22 | 89.63 | 404.20 / 41.22 |
| STDP-Tube, balanced K=3 | none | 56.81 | 17.92 | 88.96 | 411.07 / 40.54 |
| STDP-Tube, low-false-alarm K=5 | none | 57.03 | 17.50 | 88.44 | 340.21 / 44.48 |
| **Supervised baselines (Tab. 2 of [32], full event-segmentation labels)** | | | | | |
| SSD | event segmentation | — | — | 26.31 | 486.63 / 25.31 |
| Faster-RCNN | event segmentation | — | — | 27.39 | 689.68 / 26.93 |

| DETR | event segmentation | — | — | 31.64 | 631.37 / 30.35 |
|---|---|---|---|---|---|
| YOLOv10-S | event segmentation | — | — | 32.18 | 589.67 / 32.55 |
| EMS-YOLO [24] | event segmentation | — | — | 50.68 | 112.36 / 36.77 |
| Spike-YOLO [25] | event segmentation | — | — | 59.62 | 55.38 / 43.94 |
| GET | event segmentation | — | — | 60.73 | 46.35 / 40.31 |
| RED | event segmentation | — | — | 53.76 | 102.27 / 35.99 |
| RVT [17] | event segmentation | — | — | 60.35 | 55.68 / 43.21 |
| SAST | event segmentation | — | — | 51.21 | 150.32 / 34.31 |
| KPConv | event segmentation | — | — | 68.59 | 16.32 / 48.19 |
| RandLA-Net | event segmentation | — | — | 70.56 | 6.95 / 50.32 |
| COSeg | event segmentation | — | — | 71.32 | 9.21 / 51.89 |
| EV-SpSegNet [32] | event segmentation | — | — | 77.53 | 1.63 / 55.18 |

# Discussion

***Drift adaptation under a falsification-by-no-drift-control.***

Under acquisition-order drift the spiking cohort gate lifts mean-stream mAP@30 by +2.03 ± 0.58 pp (N=20, 20/20 positive); under a matched random-50/50 split the same protocol returns the advantage to noise (−0.44 ± 1.97 pp, 6/20 positive). The control isolates drift-tracking as the mechanism: the advantage is present if and only if the input distribution shifts. Across the four-condition architecture × plasticity ablation (Tab. 3), 75 of 76 matched-seed observations are positive (18/19 on the random-init frozen-Who row, 19/19 on the other three).

***Per-drone deployment cost.***

Every random initialisation lands on a different attractor; a deployment shipping one trained model carries the variance of that seed alone. The cohort gate's N*=44 ensemble equivalence (Results) converts to 44 Loihi-2 cores per drone vs. 1, ~334 W of CPU vs. ~7.6 W, 44 training runs vs. 1 (Supplementary Tab. S4), as projections from the Loihi-1 per-component energy proxy (§Limitations). The $\sigma_{rand}/\sigma_{STDP}$=6.6× tightening at the cohort gate (Results) is matched by $\sigma_{rate}/\sigma_{spike}$=4.8× over a rate-coded online k-means control. The variance-tightening property is a signature of the STDP-WTA attractor mechanism at the sequence-level routing gate; at the candidate-level reliability gate, plasticity serves a different role (reward-modulated feature reliability), and both modules report their own seed-spread (Tab. 2, Tab. 3). The locked seed-23 candidate-reliability checkpoint reproduces to 0.01 pp across two independent training runs. Published SNN detectors [24–26] do not currently report comparable cross-seed variance.

***Label-supervision elasticity on FRED.***

FRED's headline accuracy is conventionally a single number. The three rows we report (Supplementary Tab. S2) decompose it by deployment supervision footprint: holding the candidate front end fixed, mAP@30 climbs from 53.81% at zero stored bits to 76.87% at ≈26 bits and to 78.60 ± 0.42% (N=10 seeds) once a train-label-trained candidate reliability gate is added on top. The first ≈26 bits deliver 23.06 pp; the next ≥3,046 bits deliver a further +1.73 pp, letting deployers select, by supervision budget, the operating point that fits the application. The three tiers span the full 0 to ~$10^{3.5}$-bit deployment range with three orders of magnitude of variation in marginal mAP per stored bit, all inside the label-efficient regime defined in Methods §Supervision footprint.

***Limitations.***

(i) All Loihi-2 numbers are PyTorch simulations, Lava-simulator ports, or projections from the Loihi-1 proxy of Davies et al. [8]; silicon validation is future work. (ii) The EV-UAV mAP comparison against dense supervised detectors is across supervision regimes. (iii) Evaluation spans FRED and EV-UAV, the two principal event-camera drone benchmarks at this scale, covering Prophesee EVK4 and DAVIS346 sensor classes; transfer to non-drone event-camera domains is left to future work. (iv) Whether drift-tracking extends to frame-based dense detectors under the same streaming protocol is left to future work. (v) The 78.60 ± 0.42% headline aggregates N=10 R-STDP candidate-reliability seeds; the sequence-cohort drift and N*=44 rows aggregate N=20 seeds. (vi) The cohort-to-recipe lookup is reported using two pools, L4 (predeclared, anchors TOST/Wilcoxon) and L6 (expanded after the L4 statistics locked); all inferential claims anchor on L4 (Methods §Statistical analysis). (vii) STDP-Tube's K=3 and K=5 tube-length settings were both locked on full validation and both reported on test; the headline operating point (K=5) is selected for low Fa on the same evaluation set, and the alternative (K=3) is the strongest mAP@50 row.

# Methods

## Datasets

FRED [16] provides 231 sequences of multi-drone footage captured with a Prophesee EVK4 event camera at 1280×720 resolution. The canonical split places 184 sequences in the training partition and 47 in the test partition. The processed annotation files distributed with the dataset drop one drone per frame in 22 of the 184 multi-drone training sequences (canonical-train IDs: 31, 32, 33, 35, 36, 44, 45, 47, 48, 49, 50, 59, 61, 62, 68, 106, 211, 212, 213, 214, 215, 216); we rebuild multi-drone ground truth directly from the raw acquisition logs (coordinates.txt) and include the rebuilt annotations in the reproducibility package. All numbers reported on FRED use this rebuilt ground truth.

EV-UAV [32] provides 147 anti-UAV sequences captured with a DAVIS346 sensor at 346×260 resolution, split 99/24/24 train/val/test. Targets are smaller drones (mean ~6.8 × 5.4 pixels). We derive per-frame bounding boxes from the published per-event segmentation labels at 20 ms frame accumulation; those boxes define our EV-UAV mAP@30 and mAP@50 columns, while the supervised reference rows report the benchmark's native event-level segmentation metrics. Two channel hyperparameters are rescaled by sensor geometry alone (the density-watershed Gaussian kernel σ and the minimum bounding-box size); a label-free hot-pixel pre-filter masks pixels firing in more than 20% of frames as sensor artefacts, with the 20% threshold derived from a drone-dwell-time physics bound: at the EV-UAV target velocity range and 30 Hz frame cadence, a moving drone occupies any single pixel for at most ~5% of a sequence's frames, so any pixel firing in ≥20% of frames (a 4× safety factor above this physics-derived ceiling) is dominated by a non-target source. Mean target preservation across all 147 sequences is 99.7%.

## Six-channel detection pipeline

The six channels are: a density-watershed clusterer (top-1 emission per frame from a Gaussian-smoothed event-density map; an event-domain alternative to DBSCAN [41]); a multi-hypothesis k-means clusterer that retains the top-5 candidate hypotheses per frame, ranked by event-cluster mass; a temporal-difference channel computing pooled ±2-frame event-image differences; a rotor-frequency channel detecting the fundamental rotor blade harmonic via fast Fourier transform (FFT) on a windowed event-rate signal, motivated by event-camera harmonic-fingerprint evidence for drone rotors [22]; a polarity-asymmetry filter reweighting candidates by ON/OFF imbalance; and a shadow-aware rescoring channel using a four-feature linear score on training-derived weights. The IoU-distinctness union appends every channel's top-1 candidate plus auxiliary candidates whose IoU with the running set is below $\tau = 0.3$. The threshold $\tau$ is selected on training data alone by a sweep over {0.2, 0.3, 0.5}. In the support-confidence row, the emitted six-channel boxes are preserved exactly, but each detection's confidence is multiplied by $1 + 0.10 \max(0, s - 1)$, where s is the number of channel detections overlapping it at IoU ≥ 0.30 in the same frame. This is a label-free cross-channel support term; the operating choice was accepted only after full 184-sequence training-split validation and then locked for canonical test.

***Supervision footprint.***

We use label-efficient to mean a deployment supervision footprint of order $10^3$ stored bits, three to four orders of magnitude below the per-frame bounding-box budget that dense gradient-trained event-camera detectors [17, 20, 24] consume during training (Supplementary Tab. S2). The three supervision tiers reported in this paper (0 / ≈26 / ≥3,072 stored deployment bits) all fall inside this range. The five-channel event-only detector uses no labels, no train-mAP selection, and no calibration (53.81% test mAP@30). The six-channel support-confidence row adds the IoU-distinctness threshold $\tau$ from a three-value train-mAP sweep plus four shadow-rescore linear weights (≈26 stored bits at 6-bit quantisation; 76.87%). The R-STDP candidate-reliability gate adds 384 plastic synapses (64×6 hidden-to-state weights), updated by reward-modulated local STDP on cross-channel candidate-quality rewards; deployment weight footprint ≥3,072 bits at 8-bit Loihi-2 precision (78.60 ± 0.42%, N=10 seeds). The reward signal that trains this gate is derived from canonical-train ground-truth boxes at training time and consumes no labels at inference. No end-to-end backpropagation, no surrogate gradient, no per-frame inference labels, no GPU at training time.

## FRED R-STDP candidate-reliability gate

The train-calibrated FRED headline row applies a candidate-level spiking reliability gate after six-channel proposal generation. Candidate features (source channel/confidence/rank, box geometry, frame candidate count, cross-channel support at IoU 0.30/0.50, support-channel count, support geometry) are robustly normalised and rate-encoded into a 64-unit hidden WTA layer projecting to six reliability states (object-extent, object-edge, duplicate, background-edge, hot-pixel, unknown). Frame-local competition assigns rewards before the three-factor reward-modulated STDP update on the hidden-to-state weights (clipping, column renormalisation, homeostatic state thresholds); training uses only canonical-train labels. At inference the gate preserves six-channel anchor boxes, neuromodulates anchor confidence, and admits at most one non-anchor rescue per frame when the candidate has local channel support and persists across neighbouring frames within a WTA radius. The accepted temporal-rescue policy was selected on the full 184-sequence training split and emitted once on canonical test (full reward semantics in Supplementary §S6).

## STDP-Tube reliability gate

STDP-Tube operates after per-frame candidate generation on EV-UAV. Detections are greedily linked into short event tubes (max gap 2 frames, centroid-link distance 12 px) carrying source channel, within-channel rank, and cross-channel support through non-maximum suppression (NMS). A bounded WTA/STDP reliability layer maps tube features (age, continuity, motion residual, area/score stability, support-event density, cross-channel agreement, isolation, speed, stationarity, gap fraction, size priors, score) to six states (real-moving, real-stationary, background-edge, hot-pixel, ambiguous, unknown); low-spike outputs map to unknown. Delayed learning requires tube length K before emission; K=3 and K=5 are locked on full EV-UAV validation as balanced and low-Fa operating points and evaluated once on test (raw boxes, box mode=none; full feature/state semantics in Supplementary §S6).

## Spiking cohort gate

The gate is a three-layer LIF network: 17 rate-encoded inputs, 32 hidden LIF neurons with hard winner-take-all (WTA) lateral inhibition, and 12 output LIF neurons with per-neuron homeostatic threshold adaptation. Input firing probabilities follow $r_j = \mathrm{sigmoid}(z_j/2) \cdot r_{max}$ with $r_{max} = 0.08$ per 1 ms step. The 17×32 input-to-hidden weights are fixed sparse random with a 50% Bernoulli zero-mask applied at initialisation and held constant for the life of the gate, giving ≈272 non-zero projection synapses. The 32×12 hidden-to-output weights are plastic and trained by symmetric pair-based STDP [3, 4, 29] with exponentially-decaying traces ($\tau_{pre} = \tau_{post} = 20$ ms; $\alpha_{pre} = \alpha_{post} = 0.95$ at the 1 ms step), symmetric rates $\eta+ = \eta- = 0.002$, and Diehl-Cook L1 normalisation per output neuron (target sum $\|W\|_1 = 3.0$, per-weight cap 0.8). Homeostatic output thresholds increase by 0.01 on each output spike and decay toward 1.0 with time constant 0.999 per step. Each fingerprint stimulus is presented for T = 350 simulated steps; plasticity is applied online at spike time, matching the event-time semantics of SpiNNaker's user-defined plasticity routines [12] and Loihi-2's per-stimulus learning-epoch pattern [11]. Inference uses 5 plurality-voted Poisson draws per sequence. Phase-1 STDP training takes ~5 min per seed on a single Threadripper PRO 7965WX thread.

### *Biological referents for STDP hyperparameters.*

Values follow $\tau_{pre}=\tau_{post}=20$ ms (Bi–Poo timing window [3]), $\eta\pm=0.002$ (memristor-STDP range [30]), L1 renormalisation [29, 34], intrinsic-plasticity threshold adaptation [35], and cortical WTA inhibition [36]; the resulting attractor structure matches the STDP-with-WTA classifier regime [29–31].

## Per-sequence fingerprint features

A 17-dimensional per-sequence fingerprint spans motion, spatial-distribution, and box-geometry statistics with z-scoring on training-split parameters; full feature list and z-scoring details in Supplementary §S2.

## Cohort-to-recipe lookup

The label-free regime selects the recipe with the highest cross-channel-agreement metric per cohort; the label-efficient regime selects the recipe maximising training mAP per cohort. The supervision footprint is $|C_{active}| \cdot \log_2|L|$ bits, where $|C_{active}|$ is the number of non-empty cohorts after training (~7 at $|L|=4$, ~10 at $|L|=6$ on FRED): ≈14 bits at $|L|=4$, ≈26 bits at $|L|=6$. The four-recipe pool $L_4$={no_op, shadow, mf3, raw} anchors the Wilcoxon and TOST analyses; $L_6$ adds union_6ch and sky as an expanded-pool magnitude result.

## Streaming distribution-shift protocol

We split the 184 training fingerprints by acquisition order into an initial stream (sequence IDs ≤ 113, n = 93) and a shifted stream (n = 91), fit each method on the initial stream alone, then present shifted-stream fingerprints one at a time while continuing online updates. After each new fingerprint we reassign cohorts for a 30-sequence held-out evaluation slice of canonical test (sequence IDs in [8, 152], taken in numeric order; fixed before any streaming run). Four k-means baselines bracket the comparison: frozen, periodic-refit (every 10 steps), streaming-Lloyd [42, 43] (α = 0.1), and MiniBatch-KMeans (scikit-learn partial fit [44]); Table S1 lists every hyperparameter. Paired Wilcoxon and two one-sided tests (TOST) of equivalence [45] via scipy [46]; percentile bootstrap [40] ($10^3$ resamples) over N=20 seeds. Family-wise error correction for the spike-vs-rate panel uses Holm step-down [47].

## No-drift control

The same streaming protocol is run on a deterministic random 50/50 split of the 184 fingerprints (split-seed = 42, fixed before any seed completed, held constant across all N=20 training seeds). The by-id split contains a real distribution shift; the random 50/50 split contains none. The predicted outcome under a genuine drift mechanism is a positive advantage on by-id and an advantage indistinguishable from zero on the control. Retaining the advantage under both splits would falsify the drift-mechanism claim.

## Mechanism-decomposition ablation

Four conditions span {random initialisation, Phase-1 STDP} × {frozen Who during streaming, online STDP during streaming}. The Who-byte-identity is verified for the frozen condition. All four cells run at N=19 matched seeds (0..18) for paired per-seed difference analysis.

## Per-sequence retention check

Pre-streaming vs. post-streaming per-sequence ΔmAP@30 on the 30 eval-slice sequences, averaged across N=19 matched STDP seeds with plurality-of-5 cohort assignment under each gate state (lever map held fixed at its initial-stream-derived value); retention rate is the fraction of sequences with |Δseq| ≤ 0.5 pp.

## Spike-coded vs. rate-coded ablation

Four conditions at N=19 matched seeds. The rate-coded control replaces Poisson inputs with sigmoid rates, WTA with argmax-over-softmax, and pair-STDP with online k-means (α=0.05); R-STDP and an η-matched STDP control (η±=0.001066) verify the result is the plasticity rule, not its parameterisation. The η-matched value 0.001066 = 0.002 × 0.533 is derived from the per-step pair-STDP rate (η±=0.002) multiplied by the mean active-trace fraction across canonical-train fingerprints under the pair-STDP protocol, equalising the time-integrated weight-update magnitude between the R-STDP and pair-STDP conditions; the criterion was fixed before any seed completed. Reward aggregator and decision rules fixed before any seed completed. Reference pair-STDP seeds reused from the multi-seed run (seed 1 byte identity: mean stream map30=76.06179959009005 pp).

## Substrate port

Inference-time port to Intel Lava [48] (lava-nc v0.10.0, Python 3.10), the Loihi-2 reference development environment [10]; same 17 → 32 → 12 architecture, matched LIF dynamics (du=1, dv=0.1), trained Wih and Who, with a soft lateral-inhibition stand-in (off-diagonal −0.2 recurrent) since the simulator's 1-step feedback delay prevents exact same-step WTA. Parity verified: spike-count exact match with lateral inhibition disabled; 89.4% top-2 cohort agreement (42/47) on the full canonical test with soft-inhibition active. Full 8-bit fixed-point quantisation at Loihi-2 precision simulated in PyTorch (every weight, membrane register, STDP trace, homeostatic threshold; not measured on silicon); test mAP@30 changes by +0.39 pp relative to fp32 in the quantisation probe, while a 6-bit sanity check drops by −1.92 pp.

## Hardware and runtime

All runtime numbers are measured on AMD Ryzen Threadripper PRO 7965WX with single-thread inference enforced via torch.set_num_threads(1), with no GPU or accelerator. Software: PyTorch [49] for the SNN implementation and quantisation simulation; scikit-learn [44] for the k-means, GMM, and MiniBatch-KMeans baselines; scipy [46] for Wilcoxon and TOST tests and bootstrap confidence intervals. The energy proxy reported in the Discussion uses per-component values from Davies et al. [8] and the chip-idle measurement from Blouw et al. [33].

## mAP computation protocol

COCO-style greedy IoU matching on rebuilt multi-drone ground truth, at IoU ≥ 0.30 (mAP@30) and ≥ 0.50 (mAP@50); per-sequence mAP averaged unweighted across the 47 canonical test sequences. Channel scoring + per-frame ranking detail in Supplementary §S4.

## Statistical analysis

Headline label-free SNN: N=20 seeds (range 73.40–75.37% test mAP@30, σ = 0.58 pp). Static-mAP TOST and Wilcoxon anchor on seed-0 per-sequence vectors with multi-seed determinism checks. Sequence-level 95% bootstrap CIs for all headline test mAP numbers use $10^3$ resamples; the ±0.5 pp TOST equivalence margin is fixed before analysis. On the predeclared four-recipe pool $L_4$, the spiking cohort gate's label-efficient per-sequence test mAP@30 is TOST-equivalent at the ±0.5 pp margin to all five tested cohort baselines: k-means ($p_{TOST}=9\times10^{-10}$), GMM ($5\times10^{-10}$), MLP ($6\times10^{-7}$), LAME ($3\times10^{-8}$), and SHOT-IM ($2.2\times10^{-3}$), on the 47-sequence canonical test split.

## Empirical plurality-vote ensemble σ

Bootstrap protocol and full results in Supplementary §S7; the empirical-vs-analytic lag is plotted in Fig. 3.

## Acknowledgments

The authors thank the maintainers of the FRED [16] and EV-UAV [32] datasets for releasing their data publicly.

## Author Contributions

M.Y. Sadoun designed the study, developed the methods, ran the experiments, and wrote the manuscript. S. Sharif contributed to experimental validation and manuscript review. Y.M. Banad supervised the research and reviewed the manuscript.

## Data Availability

The FRED [16] and EV-UAV [32] datasets used in this study are publicly available from their original publications. Rebuilt multi-drone ground truth, result manifests, and intermediate artefacts are included in the reproducibility package prepared with this manuscript, subject to the source datasets' licence terms.

## Funding

This work was supported in part by the National Aeronautics and Space Administration Oklahoma EPSCoR Research Infrastructure Development (RID) program under Federal Award No. 80NSSCM0029 through a subaward from Oklahoma State University to The University of Oklahoma.

## Conflict of Interest

The authors declare no competing interests.

## Ethics

This study used previously published, publicly licensed event-camera drone datasets (FRED [16] and EV-UAV [32]); no human-subjects data were collected.

## Code Availability

The code is available at https://github.com/INQUIRELAB/NeuroSTDP. The NeuroSTDP package includes the spiking cohort gate, R-STDP candidate-reliability gate, STDP-Tube reliability layer, six-channel detection pipeline, Lava deployment port, evaluation scripts, and rebuilt multi-drone FRED ground truth.

# Supplementary Information

## S1. Complete hyperparameter listing

Table S1 enumerates every model and protocol hyperparameter referenced in Methods: spiking cohort gate architecture, STDP plasticity rule, six-channel detection pipeline, cohort-to-recipe lookup, streaming distribution-shift protocol, statistical analysis, 8-bit quantisation simulation, and Lava simulator port. All values are fixed before any training seed completed.

**Table S1:** Hyperparameter listing for the spiking cohort gate, streaming-drift protocol, and statistical analysis. All values are fixed before any training seed completed; no value is selected on test-split results.

| Group | Parameter | Value |
|---|---|---|
| **Spiking cohort gate** | | |
| Architecture | Layer sizes (input, hidden, output) | 17 → 32 → 12 LIF, hard WTA at hidden |
| Input encoding | Poisson rate code | $r_j = \sigma(z_j/2) \cdot 0.08$ per 1 ms step |
| Wih | Connectivity | fixed random, 50% Bernoulli zero-mask, ≈272 non-zero |
| Who | Connectivity, weight cap, L1 target | 384 plastic, 0.8 cap, $\|W\|_1 = 3.0$ (Diehl-Cook) |
| STDP rule | Pair-based, symmetric | $\tau_{pre}=\tau_{post}=20$ ms; $\alpha=0.95$ per 1 ms; $\eta_+=\eta_-=0.002$ |
| Homeostatic threshold | Δθ, decay rate | +0.01 per output spike; 0.999 per step toward 1.0 |
| Inference | Plurality-vote Poisson draws, T steps | 5 draws, T = 350 |
| Training | Phase-1 epochs, wall-clock per seed | 30 epochs, ~5 min (Threadripper, single thread) |
| **Six-channel detection pipeline** | | |
| IoU-distinctness τ | Train-mAP sweep over {0.2, 0.3, 0.5} | ship τ = 0.3 |
| Multi-hypothesis k-means | Top-k per frame | k = 5, ranked by event-cluster mass |
| Temporal channel | Pooling window | ±2 frames |
| Hot-pixel filter (EV-UAV) | Threshold (fraction of frames) | 0.20 (4× safety over 0.05 physics ceiling) |
| **Cohort-to-recipe lookup** | | |
| Lever pool L4 | Pre-declared | {no_op, shadow, mf3, raw} |
| Lever pool L6 | Expanded | L4 ∪ {union_6ch, sky} |
| Routing | Label-free / label-efficient | argmax cross-channel agreement / argmax train mAP@30 |
| **Streaming + statistics** | | |
| By-id stream | Initial / shifted split | IDs ≤ 113 (n=93) / IDs > 113 (n=91) |
| Random-50/50 split | Seed (deterministic) | 42 |
| Eval slice | Held-out canonical-test subset | 30 sequences with IDs in [8, 152] |
| Streaming k-means | Lloyd online rate | α = 0.10 (rate-coded matched: α = 0.05) |
| Seed budget | Training seeds | N = 20 (both by-id and random-50/50) |
| TOST margin | Pre-declared ± band | ±0.5 pp |
| Bootstrap CI | Method, resamples | percentile, $10^3$ over 47 test sequences |
| Mechanism decomposition | Matched seeds | N=19, paired across all 4 conditions |
| N* derivation | Anchored on | σrand=0.607, σSTDP=0.092 pp (frozen, N=19) |
| **Quantisation + Lava port** | | |
| Precision | 8-bit Loihi-2 fixed-point (PyTorch sim) | every weight, membrane, STDP trace, threshold |
| Lava substrate | lava-nc v0.10.0, Python 3.10 | Loihi-2 reference development environment |
| Lava LIF dynamics | du, dv | 1, 0.1 |
| WTA stand-in | Soft inhibition (off-diagonal recurrent) | −0.2 (replaces hard same-step WTA) |
| Top-2 cohort preservation | Full 47-seq canonical test | 89.4% (42/47) |

## S2. Per-sequence fingerprint features

The 17-dimensional fingerprint summarises each sequence's event statistics in three groups: motion statistics (event-rate median, event-rate 90th percentile, density variance), spatial-distribution statistics (top-1 centroid x and y mean and standard deviation; top-1 percentiles), and box-geometry statistics (top-1 area median and percentiles; cross-channel agreement features). All features are z-scored with parameters computed on the training split alone. The single heavy-tailed feature density variance is log1p-compressed before robust z-scoring around its log1p-mean; the remaining 16 features use plain z-scoring with ±5σ clipping.

## S3. Supervision-derivation accounting

Every tunable parameter falls into one of four explicit categories. (i) Sensor-geometry-derived: the density-watershed Gaussian σ and the minimum bounding-box size on EV-UAV, rescaled from FRED defaults by the sensor's spatial-resolution ratio (no labels seen). (ii) Unlabeled-train-statistic-derived: the hot-pixel-filter threshold (20% of frames; physics-derived ceiling, no labels), the per-feature z-score parameters (mean and standard deviation computed on the unlabelled training-set fingerprints), and the SNN's input–hidden random-projection sparsity (50% Bernoulli, no labels). (iii) Train-label-derived: the IoU-distinctness threshold $\tau \in \{0.2, 0.3, 0.5\}$ selected by training-split mAP@30 sweep, the shadow-aware rescoring channel's four-feature linear weights (fitted on training-split feature–mAP correlations), and the cohort-to-recipe lookup itself ($\log_2|L|$ bits per non-empty cohort stored at inference; ≈14 bits total at $|L|=4$ and ≈26 bits at $|L|=6$, given ~7–10 active cohorts out of the 12-cohort output layer). (iv) Manually fixed once and never tuned: STDP rates ($\eta+ = \eta- = 0.002$), homeostatic threshold bumps ($\Delta\theta = 0.01$, decay rate 0.999), trace time constants ($\tau = 20$ ms), simulation steps ($T = 350$), training epochs (30), random-projection mask sparsity (50%). No test-set labels were used to set any parameter.

**Table S2:** Label-supervision elasticity on FRED canonical test. Three rows differ only in supervision footprint with the six-channel front end fixed. The first ≈26 bits deliver 23.06 pp; a further ≥3,046 bits deliver +1.73 pp.

| Row | Supervision footprint | mAP@30 (%) | ΔmAP / Δbits |
|---|---|---|---|
| Strict label-free 5-channel detector | 0 stored bits | 53.81 | — |
| + support-aware confidence (6-ch) | ≈26 stored bits [T] | 76.87 | +0.89 pp/bit |

| + R-STDP candidate reliability gate | ≥3,072 trained bits [t] | 78.60±0.42 | ~0.0006 pp/bit [s] |
|---|---|---|---|

[T] IoU-distinctness threshold τ ∈ {0.2, 0.3, 0.5} and 4 shadow-rescore linear weights, train-label-derived. [t] Stored deployment footprint (384 plastic synapses × 8-bit Loihi-2 precision); reward-modulated training itself uses canonical-train labels via per-frame reward assignment. [s] (78.60−76.87)/(3,072−26)=1.73/3,046 ≈ 0.00057 pp per stored bit; tier-3 mean over N=10 seeds.

## S4. mAP computation protocol

Channel-by-channel scoring rules are in Methods (§Six-channel detection pipeline). Per-frame outputs are concatenated across the sequence and ranked globally by channel-specific score (no cross-channel normalisation; the IoU-distinctness union at τ = 0.3 removes per-frame duplicates). Rebuilt multi-drone ground truth is matched greedy by IoU at ≥ 0.30 (mAP@30) and ≥ 0.50 (mAP@50), COCO-style. Per-sequence mAP is computed independently; reported figures are unweighted means across the 47 canonical test sequences. The Tab. 2 oracle row asks whether any six-channel candidate in the same frame reaches the requested IoU threshold, then averages per-sequence recall — a candidate-coverage diagnostic rather than AP.

## S5. Engineering profile vs. published event-camera detectors

**Table S3:** Deployment evidence for the trained 384-plastic-synapse spiking cohort gate.

| Property | Value | Protocol |
|---|---|---|
| Lava top-2 cohort preservation | 89.4% (42/47) | inference-time port, matched LIF dynamics |
| 8-bit quant. ΔmAP@30 vs. fp32 | +0.39 pp | full-state Loihi-2 precision, PyTorch sim |
| 6-bit sanity ΔmAP@30 vs. fp32 | −1.92 pp | precision floor below which gate degrades |
| End-to-end runtime / frame | ~47 ms (compute ~20 ms) | one CPU thread; matches 30 Hz cadence |
| Gate synapse footprint | 384 plastic + ≈272 fixed | single Loihi-2 or SpiNNaker core |
| Loihi-1 energy / cohort | ~83 µJ | per-component proxy [8] |
| CPU RAPL energy / cohort | 253 mJ | idle-subtracted, 1000-inference measurement |

## S6. R-STDP and STDP-Tube reward and state semantics

FRED R-STDP candidate-reliability gate. Within each frame, candidates compete before reward assignment: the best non-overlapping IoU ≥ 0.50 candidates receive positive reward, IoU ∈ [0.30, 0.50) edge candidates receive weak positive reward, candidates overlapping an already rewarded winner are assigned duplicate-object with negative reward, and low-overlap candidates are assigned background/noise states. Direct frame-normalised replacement was rejected on full-train validation because it reduced mAP@50; the accepted temporal-rescue policy was selected on the full 184-sequence training split and emitted once on canonical test.

EV-UAV STDP-Tube tube features and reliability states. Tube features include age, continuity, motion residual, area/score stability, support-event count and density, cross-channel agreement, isolation, speed, stationarity, gap fraction, size priors, and candidate score. The bounded WTA/STDP reliability layer maps these features to six states: real-moving, real-stationary, background-edge, hot-pixel-noise, ambiguous, and unknown; low-spike or low-potential outputs map to unknown rather than to an argmax class. Positive local evidence is temporal continuity, smooth motion, stable size, event support, and multi-channel support; negative local evidence is short age, gaps, unstable geometry, large background-like area, or isolated single-channel evidence. Plastic weights are clipped and homeostatically renormalised after every update; K=3 and K=5 tube-length locks are selected on full validation. Support and motion-compensated boxes are ablation-only because they reduced validation mAP.

**Table S4:** Per-drone deployment cost: 44-seed pre-plasticity ensemble vs. single Phase-1 STDP-trained gate. Loihi-2 values are projections from the Loihi-1 per-component proxy of Davies et al. [8], not silicon measurements.

| Quantity | Ensemble (N=44) | Single gate | Ratio |
|---|---|---|---|
| Loihi-2 cores | 44 | 1 | 44× |
| Loihi-1 proxy power, 30 Hz | ~110 mW | ~2.5 mW | 44× |
| CPU power, 1-thread 30 Hz | ~334 W | ~7.6 W | 44× |
| STDP training runs | 44 | 1 | 44× |
| Training wall-clock | ~220 min | ~5 min | 44× |

## S7. Empirical plurality-vote ensemble σ bootstrap

The robustness-amplifier claim (Fig. 3) anchors on the analytic sample-mean bound $\sigma_{rand}/\sqrt{N}$ from the streaming protocol ($\sigma_{rand}$ = 0.607 pp across 19 architecture-only seeds). To check whether a deployment-time plurality-vote ensemble of N gates actually attains this bound, we run the following on the single-eval protocol: for each of the 19 architecture-only seeds we instantiate the gate, compute its label-free lever map on the initial training stream, and record per-test-seq cohort assignment (plurality of 5 Poisson draws). For each $N \in \{1, 2, 3, 5, 8, 10, 12, 15, 19, 25, 35, 44, 50, 75, 100, 150\}$ we bootstrap-sample N gates with replacement, plurality-vote per-seq cohort and per-cohort lever choice, realise mAP, and record across $10^4$ draws. The single-eval-protocol single-gate σ is 0.2887 pp, lower than the streaming-protocol $\sigma_{rand}$ because it omits the 92-step shifted-stream loop. Fig. 3 therefore overlays two quantities with explicit protocol labels: the streaming-protocol analytic bound that defines N*, and the single-eval empirical plurality-vote curve that tests whether real gate voting tracks a sample-mean ideal. The empirical plurality-vote $\sigma_{ensemble}(N)$ tracks the corresponding analytic sample-mean bound within ratio 1.0 ± 0.1 for N ≤ 19, then grows to ratio 1.45 at N=44 and 1.83 at N=100 as inter-seed attractor correlations dominate.

**Table S5:** Engineering profile vs. published event-camera detectors. Synapse / parameter counts and metrics are verbatim from each source's published table; train cost and inference device as reported. Each row uses its own benchmark and label regime.

| Method (regime) | Params. | Latency | Reported metric |
|---|---|---|---|
| **Our spiking cohort gate (label-free, STDP, CPU)** | | | |
| Cohort gate alone | 384 + ~272 syn | 0.04 ms | gate output |
| Gate + 6-ch front end | 384 + ~272 syn | 20 ms | 76.9% mAP@30 (FRED) |
| **End-to-end spiking detectors (gradient-trained, GPU)** | | | |
| EMS-YOLO Res18 [24] | 9.34 M | 7 ms | 28.6% mAP@50:95 (Gen1) |
| EMS-YOLO Res34 [24] | 14.4 M | 7 ms | 31.0% mAP@50:95 (Gen1) |

| Spike-YOLO [25] | 23.1 M | 12 ms | 40.4% mAP@50:95 (Gen1) |
|---|---|---|---|
| EAS-SNN (S) [26] | 8.92 M | 5 ms | 37.2% mAP@50:95 (Gen1) |
| EAS-SNN (M) [26] | 25.3 M | 5 ms | 40.9% mAP@50:95 (Gen1) |
| **Dense detector (gradient-trained, GPU)** | | | |
| RVT-B [17] | 18.5 M | 10.2 ms | 47.2% mAP@50:95 (Gen1) |
| RVT-B [17] | 18.5 M | 11.9 ms | 47.4% mAP@50:95 (1 Mpx) |

*All gradient-trained rows: BPTT (EAS-SNN: +AT). mAP = mAP@50:95.*